# CascadeTabNet: An approach for end to end table detection and structure recognition from image-based documents

Devashish Prasad, Ayan Gadpal, Kshitij Kapadni, Manish Visave, Kavita Sultanpure
Pune Institute of Computer Technology, India
{devashishkprasad,ayangadpal2,kshitij.kapadni,manishvisave149}@gmail.com
kasultanpure@pict.edu

## Abstract

*An automatic table recognition method for interpretation of tabular data in document images majorly involves solving two problems of table detection and table structure recognition. The prior work involved solving both problems independently using two separate approaches. More recent works signify the use of deep learning-based solutions while also attempting to design an end to end solution. In this paper, we present an improved deep learning-based end to end approach for solving both problems of table detection and structure recognition using a single Convolution Neural Network (CNN) model. We propose CascadeTabNet: a Cascade mask Region-based CNN High-Resolution Network (Cascade mask R-CNN HRNet) based model that detects the regions of tables and recognizes the structural body cells from the detected tables at the same time. We evaluate our results on ICDAR 2013, ICDAR 2019 and TableBank public datasets. We achieved 3rd rank in ICDAR 2019 post-competition results for table detection while attaining the best accuracy results for the ICDAR 2013 and TableBank dataset. We also attain the highest accuracy results on the ICDAR 2019 table structure recognition dataset. Additionally, we demonstrate effective transfer learning and image augmentation techniques that enable CNNs to achieve very accurate table detection results. Code and dataset has been made available at:* `https://github.com/DevashishPrasad/CascadeTabNet`

## 1. Introduction

The world is changing and going digital. The use of digitized documents instead of physical paper-based documents is growing rapidly. These documents contain a variety of table-based information with variations in appearance and layouts. An automatic table information extraction method involves two subtasks of table detection and table structure recognition. In table detection, the region of the image that contains the table is identified while table structure recognition involves identification of the rows and columns to identify individual table cells. The prior proposed approaches solved these two sub-problems independently.

In this paper, we propose CascadeTabNet, an improved deep learning-based end to end approach for solving the two sub-problems using a single model. The problem of table detection is solved using instance segmentation. We perform table segmentation on each image where we try to identify each instance of the table within the image at the pixel level. Similarly, we perform table cell segmentation on each image to predict segmented regions of table cells within each table to identify the structure of the table. Table and cell regions are predicted in a single inference (at the same time) by the model. Simultaneously, the model classifies tables into two types as bordered (ruling-based) and borderless (no ruling-based) tables. The model predicts the segmentation of cells only for the unbordered tables. We use a simple rule-based conventional text detection and line detection algorithms for extracting cells from bordered tables.

We demonstrate the effectiveness of iterative transfer learning to make the CNN learn from less amount of training data as well as enable it to perform well on multiple datasets by fine-tuning it on respective datasets. A new way of image augmentation was also implanted into the training process to enhance the accuracy of table detection and helping it learn more effectively.

Evaluation for table detection task was performed on three public datasets of the ICDAR 2013, ICDAR 2019 competition (Track A) dataset and TableBank dataset. We achieve 3rd rank in post-competition results of ICDAR 2019 for table detection. We achieve the highest accuracy for table detection task on ICDAR 2013 dataset and all of the three subsets of the TableBank dataset. For table structure recognition tasks we evaluate the model on ICDAR 2019 dataset (Track B2) and achieve the highest rank in post-competition results.

Our main contributions made in this paper are as per the following:

1. We propose CascadeTabNet: an end-to-end deep-learning-based approach that uses the Cascade Mask R-CNN HRNet model for both table detection and structure recognition.

2. We show that the proposed image transformation techniques for image augmentation for training enhances the table detection accuracy significantly.

3. We perform a comparative analysis of various CNN models for the table detection task in which the Cascade Mask R-CNN HRNet model outperforms other models.

4. We demonstrate an effective iterative transfer learning-based methodology that helps the model to perform well on different types of datasets using a small amount of training data.

5. We manually annotated some of the ICDAR 19 dataset images for table cell detection in borderless tables while also categorizing tables into two classes (bordered and borderless) and will be releasing the annotations to the community.

## 2. Related work

In 1997, P. Pyreddy and, W. B. Croft [19] was the first to propose an approach of detecting tables using heuristics like a Character Alignment, holes and gaps. To improve accuracy, Wonkyo Seo *et al*. [22] used the Junctions (intersection of the horizontal and vertical line) detection with some post-processing. T. Kasar *et al*. [15] also used the junction detection, but instead of heuristics, they passed the junction information to SVM.

With the ascent of Deep Learning and object detection, Azka Gilani *et al*. [9] was the first to propose a Deep learning-based approach for Table Detection by using Faster R-CNN based Model. They also attempted to improve the accuracy of models by introducing distance-based augmentation to detect tables. Some approaches tried to utilize the semantic information, Such as S. Arif and F. Shafait [1] attempted to improve the accuracy of Faster R-CNN by using semantic color-coding of text and Dafang He *et al*. [12], used FCN for semantic page segmentation with an end verification network is to determine whether the segmented part is the table or not.

In 1998, Kieninger and Dengel [16], proposed the initial approach for Table Structure Recognition by clubbing the text into chunks and dividing those chunks into cells based on the column border. Tables have many basic objects such as lines and characters. Waleed Farrukh *et al*. [7], used a bottom-up heuristic-based approach on these basic objects to construct the cells. Zewen, Chi *et al*. [5] proposed a graph-based approach for table structure recognition in which they used the SciTSR dataset constructed by themselves for training the GraphTSR model.

Sebastian Schreiber *et al*. [21] were the first to perform table detection and structure recognition together with a 2 fold system which Faster RCNN for table detection and, Subsequently, deep learning-based semantic segmentation for table structure recognition. To make the model more generalize, Mohammad Mohsin *et al*. [20] used a combination of GAN based architecture for table detection and SegNet based encoder-decoder architecture for table structure segmentation.

Recently, Shubham Paliwal *et al*. [18], was first to propose a deep learning-based end-to-end approach to perform table detection and column detection using encoder-decoder with the VGG-19 as a base semantic segmentation method, where the encoder is the same and decoder is different for both tasks. After detection results for the table are obtained from the model, the rows are extracted from the table region using a semantic rule-based method. This approach uses a Tesseract OCR engine for text location.

## 3. CascadeTabNet: The presented approach

We try to focus on using a small amount of data effectively to achieve high accuracy results. Working towards this goal, our primary strategy includes :

1. Using a relatively complex but efficient CNN architecture that attains high accuracy on object detection and segmentation benchmarking datasets as the main component in the approach.

2. Using an iterative transfer learning approach to train the CNN model gradually, starting from more general tasks and going towards more specific tasks. Performing iterations of transfer learning multiple times to extract the needful knowledge effectively from a small amount of data.

3. Strengthening the learning process by applying image transformation techniques to training images for data augmentation.

We elaborate on the strategies in the following subsections and explain the pipeline of the approach.

### 3.1. Model architecture

To attain very high accuracy results we use a model that was made by the combination of two approaches. Cascade RCNN was originally proposed by Cai and Vasconcelos [2] to solve the paradox of high-quality detection in CNNs by introducing a multi-stage model. And a modified HRNet was proposed by Jingdong Wang *et al*. [25] to attain reliable high-resolution representations and multi-level representations for semantic segmentation as well as for object detection. Our experiments and analysis show that the cascaded multi-staged model with the HRNet backbone network yields the best results due to the ability of both the approaches to strive for high accuracy object segmentation.

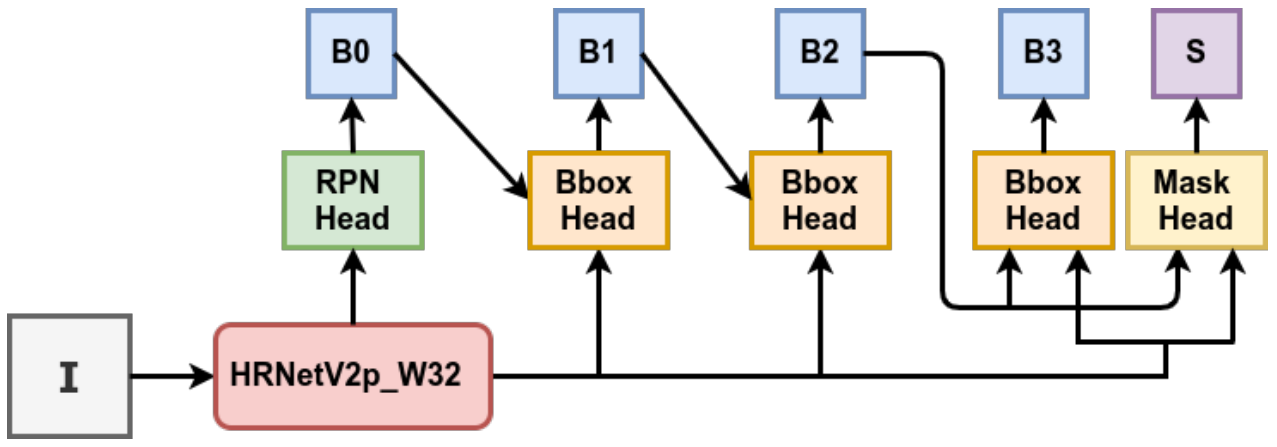

Figure 1: CascadeTabNet model architecture

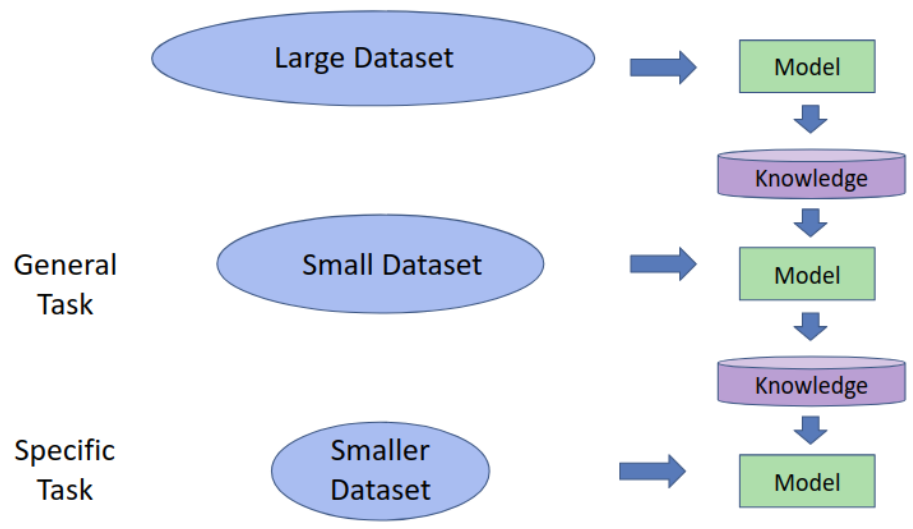

Figure 2: Two stage transfer learning

The original architecture of HRNet [14] (HRNetV1) was enhanced for semantic segmentation to form HRNetV2 [25]. And, then a feature pyramid was formed over HRNetV2 for object detection to form HRNetV2p [25]. CascadeTabNet is a three-staged Cascade mask R-CNN HRNet model. A backbone such as a ResNet-50 without the last fully connected layer is a part of the model that transforms an image to feature maps. CascadeTabNet uses HRNetV2p_W32 [25] (32 indicates the width of the high-resolution convolution) as the backbone for the model.

The architecture strategy of the Cascade mask R-CNN [3] is very similar to the Cascade R-CNN [2]. The Cascade R-CNN architecture is extended to the instance segmentation task, by attaching a segmentation branch as done in the Mask R-CNN [13]. To explain the model architecture we try to use the naming conventions similar to that of the Mmdetection framework [4]. As shown in figure 1, the image ”I” is fed into the model. The backbone CNN HR_NetV2p_W32 transforms the image ”I” to feature maps. The ”RPN Head” (Dense Head) predicts the preliminary object proposals for these feature maps. The ”Bbox Heads” take RoI features as input and make RoI-wise predictions. Each head makes two predictions as bounding box classification scores and box regression points. ”B” denotes the bounding boxes predicted by the heads and, for simplicity, we do not show the classification scores in the figure. The ”Mask Head” predicts the masks for the objects and ”S” denotes a segmentation output. At the inference, object detections made by ”Bbox Heads” are complemented with segmentation masks made by ”Mask Head”, for all detected objects.

For image segmentation using the Cascade R-CNN, Cai and Vasconcelos [3] propose multiple strategies in which the segmentation branch is placed at various stages of the network. CascadeTabNet utilizes the strategy of adding the segmentation branch at the last stage of the Cascade R-CNN. The model was implemented using the MMdetection toolbox [4]. We use the default implementation (cascade_mask_rcnn_hrnetv2p_w32_20e) of the model for our experiments and analysis.

## 3.2. Iterative transfer learning

Both the tasks involve object segmentation, and we use a multi-task learning approach as well as multiple iterations of transfer-learning to achieve our goal. In short, we first train our model on a general dataset and then fine-tune it multiple times for specific datasets. More precisely, we use two iterations of transfer learning and so we call this approach as two-stage transfer learning.

First, we create a general dataset for a general task of table detection. We add images of different types of documents like word and latex in this dataset. These documents contain tables of various types like bordered, semi-bordered and borderless. A bordered table is one for which an algorithm can use just the line positions to estimate the cells and overall structure of the table. If some of the lines are missing, it becomes difficult for a line detection based algorithm to separate the adjacent cells of the table. We call such a table as a semi-bordered table, in which some lines are not present. And a borderless table is one which doesn't have any lines. Detecting only the tables in images is a general task for an algorithm, but detecting them according to their types is a specific task. For example, detecting dogs in images is a general task, but detecting only the bulldogs and pugs is a more specific task that requires relatively more data by the model. To make it a general task for table recognition, initially, all these tables in the images are annotated as of one class (the table class), which enables the model to learn common and general features to detect tables. The trained model can use this knowledge to learn even more specific tasks like table detection according to their types.

The two-stage transfer learning strategy is used to make a single model learn end to end table recognition using a small amount of data. In this strategy, transfer learning is practiced two times on the same model. Detecting tables in images becomes a specific task for a CNN model that was earlier trained on a dataset with hundred-thousands of images to detect objects from thousand classes. So in the first iteration of transfer learning, we initialize our CNN model with the pre-trained imagenet coco model weights before training. It enables the CNN model to learn only task-specific higher-level features while getting some advantages like the lesser need for training data and reducing total training time due to beforehand knowledge. After training, CNN successfully predicts the table detection masks for tables in the images. Similarly, in the second iteration, the model is again fine-tuned on a smaller dataset to accomplish even more specific task of predicting the cell masks in borderless tables along with detecting tables according to their types. Another challenging and specific task can be table detection for a particular type of document images (latex documents). We do not freeze any of the layers in the model at any stage while performing iterative transfer learning.

For the task of table structure recognition, which involves predicting the cell masks in borderless tables along with detecting the different types of tables, we create a smaller dataset. It contains lesser images than that for table detection. This new dataset contains slightly advanced annotations intimating the model to detect tables of two types with their labels (two classes) as bordered and borderless, as well as predict borderless table cell masks (total three classes). We put borderless and semi-bordered tables in one class, the borderless class. We put semi-bordered tables in borderless class because we cannot use only line information to extract cells out of it. We need cell predictions for semi-bordered tables from the model. After again fine-tuning the model on this dataset, it successfully detects tables with their type and also predicts segmentation masks for table body and cells for borderless tables with very high accuracy.

This strategy worked effectively because while doing the knowledge transfer between two tasks the domains of both the tasks were the same. If domains of two tasks are different, for example, training a model to detect dogs in images and then using the same model to detect different types of horses, then it may result in a negative transfer. Figure 2 shows the figurative explanation to two-staged transfer learning where the same model is trained iteratively from general to a more specific task, reducing the size of the dataset as we move down.

## 3.3. Image Transformation and data augmentation

Providing a large amount of training data can easily produce deep-learning-based models that can attain very high accuracy results. Adding more training data also prevents models from over-fitting to the training data. For this concern, we try to implement image-augmentation techniques on the original training images to increase the size of training data. But, not all of these techniques would be very effective for augmenting document images. For example, the use of shear and rotation transformations won't be an effective strategy because the digital documents in the datasets are perfectly axis-aligned. We try to implement the techniques that will help the model to learn more accurately.

Documents have text or content regions and blank spaces in them. As the text elements are very small in documents and the proposed model was used for detecting real-world objects in images, we try to make the contents better understandable to the object segmentation model by thickening the text regions and reducing the regions of the blank space. We propose image transformation techniques that help the model to learn more efficiently. The transformed images are added in the original dataset, which also increases the amount of relevant training data for the model.

We propose two types of image transformation techniques as Dilation transform and Smudge transform.

### 3.3.1 Dilation transform

In the dilation transform, we transform the original image to thicken the black pixel regions. We convert the original images into binary images before applying the dilation transform. Figure 3, a) is the original image and b) is the transformed dilated image. A 2x2 kernel filter for one iteration was applied to the binary image to generate the transformed image. Experiments showed that the kernel size of 2x2 gave better results.

### 3.3.2 Smudge transform

In the smudge transform, we transform the original image to spread the black pixel regions and make it look like a kind of smeary blurred black pixel region. The original images are converted into binary images before the smudge transform is applied. In Figure 3, a) is the original image and c) is the transformed smudged image. Smudge transform is implemented using various distance transforms. The original algorithm is described by Gilani *et al.* [9] that applies Euclidean Distance Transform, Linear Distance Transform, and Max Distance Transform to the image. Also, some additional normalization and parameter tuning enhanced the results.

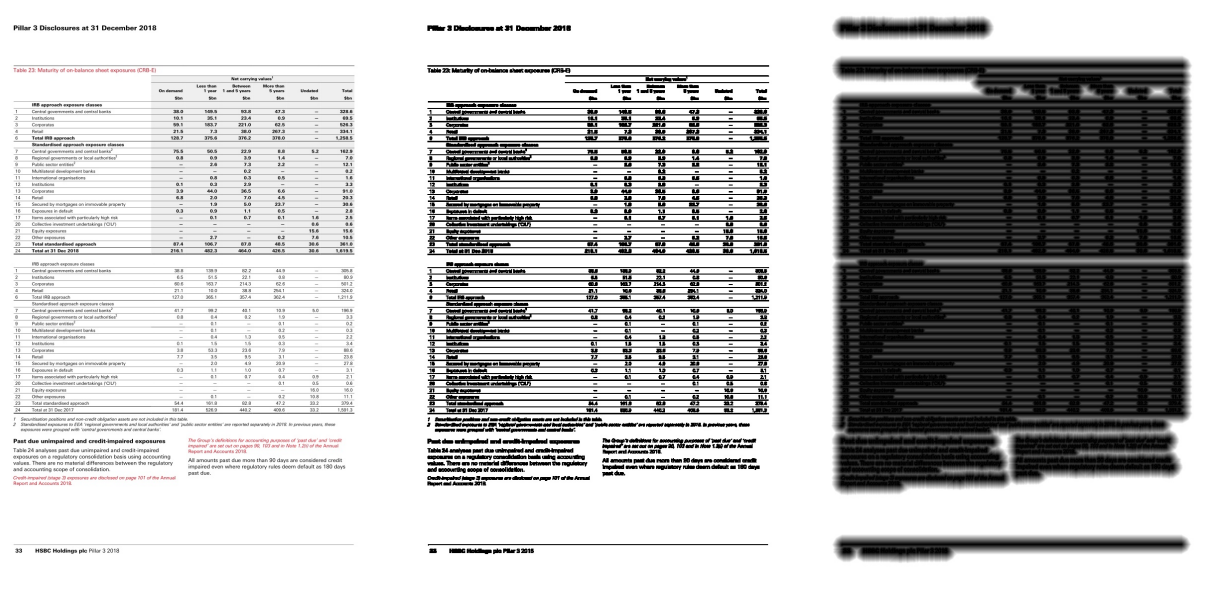

(a) Original Image (b) Dilated image (c) Smudged image

Figure 3: Image transformations

### 3.4. Pipeline

In this section, we describe various stages in the pipeline of the CascadeTabNet end to end system for table recognition.

Figure 4, shows the block diagram of the pipeline. The two-stage fine-tuned CasacdeTabNet model takes in the image of the document containing zero or more tables. It predicts the segmentation masks for tables of two types as bordered and borderless, as discussed earlier. Next in the pipeline, we have separate branches for bordered and borderless tables. Depending on the type of the detected table it is further processed by its respective branch post-processing module. Post-processing modules perform trivial tasks of arranging and cleaning the outputs of the model.

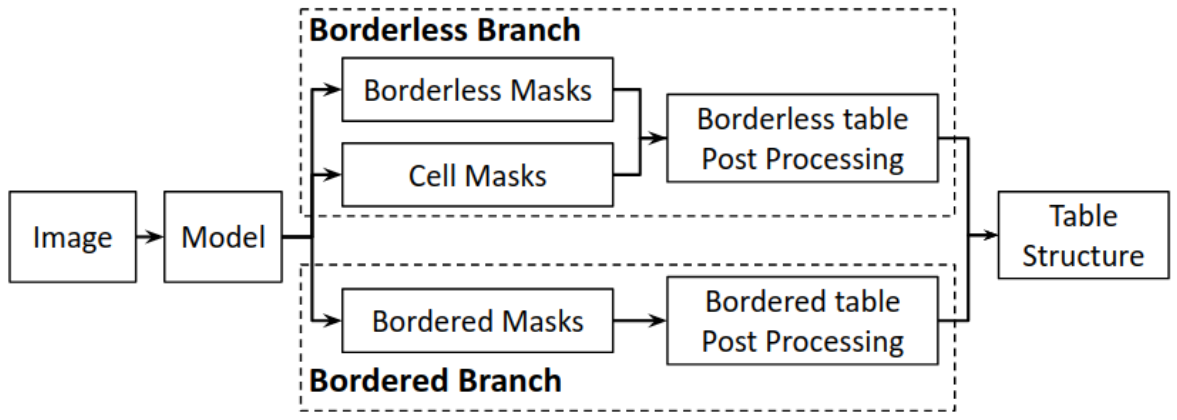

Figure 4: CascadeTabNet Pipeline

In the borderless branch, we arrange the predicted cells detected inside the table into rows and columns based on their positions. We estimate the missing table lines using the positions of identified rows and columns. Based on these lines, for undetected cells, we detect cells using a contour-based text detection algorithm. And finally, Row-span and Col-span cells are also identified after estimating the lines.

In the bordered branch, a conventional algorithm of line detection is used to detect lines of bordered tables. The cells are identified using the line intersection points. And within each cell, the text regions are detected by using the contour-based text detection algorithm. We prefer not to train our model for bordered table cell segmentation masks prediction because using the line information from bordered tables is much easier and efficient to recognize the cells.

## 4. Dataset Preparation

For creating a General dataset for table detection task we merge three datasets of ICDAR 19 (cTDaR)[8], Marmot [6] and Github [1] [23]. The cTDaR competition aims at benchmarking state-of-the-art table detection (TRACK A) containing two subsets of the dataset as Modern and Archival, further described in [8]. We include only the modern subset of this dataset in the general dataset. This subset contains several images of word and latex documents, having text in English and Chinese languages. We also include a publicly available Marmot dataset published by the Institute of Computer Science and Technology of Peking University, further described in [6]. Marmot dataset holds two subsets as Chinese and English, we include both sets in the general dataset. As done by DeepDeSRT [21], to achieve the best possible results, we removed the errors in the ground-truth annotations of the dataset. And finally, we also include a dataset from the internet [23] in the general dataset that contains only borderless table images with some magazine and newspaper based document images. This dataset was also cleaned like the marmot dataset. The General dataset contains a total of 1934 images having 2835 tables in it, and we use this dataset to train a General model.

For the preliminary analysis of image augmentation, we created four training sets. The first set contains the original images. The second set is created by applying the dilate-transform to all the images in the original set and adding them in the set along with corresponding original images. Similarly, the third set is created by applying the smudge-transform to these original images. And the last set is created by adding the smudged, dilated and original images altogether in the set. In Section 5. we perform a rigorous analysis of these training sets by training different types of models. We show the effectiveness of augmentation techniques, as it boosts the models' performance.

To evaluate the model on the ICDAR 19 (Track A Modern) competition dataset, we perform the dilate image transform for all the images of the Track A Modern dataset. And then fine-tune the General model on it.

For testing all of the aforementioned datasets, we use the test set of the ICDAR 19 dataset (Track A Modern). We find this set robust and ideal for testing because it contains all types of images like Latex and Word, having all types of tables.

We also provide evaluation results on the TableBank dataset [17]. TableBank dataset is a new image-based table detection and recognition dataset that contains Word and Latex documents based 417K table images. The table detection subset of the dataset has 163,417 images in Word, 253,817 images in Latex and 417,234 images in Word+Latex subsets respectively. To demonstrate the effectiveness of our approach, we don't fine-tune the model on the whole dataset. Instead, we fine-tune the model on a very small subset of the actual TableBank datasets. For latex, we only choose 1500 images randomly from the TableBank Latex for training. For creating the test set for latex, we randomly choose 1000 images from the TableBank Latex dataset, as originally done by the authors [17]. Similarly, for Word, we choose 1500 images randomly from the TableBank Word dataset for training. And again, for creating the test set, we randomly choose 1000 images from the TableBank Word dataset. We found that some annotations provided for the TableBank Word dataset images were inappropriate. We preferred not to include these images in the test set. And finally, we create a set for both latex and word by combining the randomly chosen images of word and latex train sets, putting a total number of 3000 images for training. And likewise, for testing, we create the test set by combining the randomly chosen images of test sets of latex and word, putting a total number of 2000 images.

[1] https://github.com/sgrpanchal31/table-detection-dataset

And we also evaluate the model on the ICDAR 13 [11] dataset that includes a total of 150 tables. It has two subsets as EU and US, in which there are 75 tables in 27 PDFs from the EU set and 75 tables in 40 PDFs from the US Government. We convert all of these PDFs into images and we get 238 images, out of which we use 40 randomly choose images for fine-tuning and others for testing.

For creating a dataset for table structure recognition task we manually annotated some images from the ICDAR 19 (Track A Modern) train set. As discussed earlier, this dataset is annotated for three classes. We randomly choose 342 images out of 600 images of the ICDAR 19 train set. It had 114 bordered tables, 429 borderless tables and 24920 cells in borderless tables in these images and were annotated accordingly. We release this dataset to the research community. The test set for table structure recognition was provided by the cTDaR competition track B2. It contains 100 images of all types of documents and tables.

## 5. Results and Analysis

In this section, we start by demonstrating the effectiveness of image transformation techniques by performing experiments with a baseline model. Then we show a comparative analysis of various CNN models with Cascade mask RCNN HRNet. And finally, we show the evaluation benchmarks of our model on public datasets. The experiments were performed on Google Colaboratory platform with P100 PCIE GPU of 16 GB GPU memory, Intel(R) Xeon(R) CPU @ 2.30GHz and 12.72 GB of RAM.

### 5.1. Preliminary Analysis

To show the effectiveness of the proposed image transformation techniques, we train a baseline model on all four datasets (created by augmenting the general dataset in section 4) and evaluate the results on ICDAR 19 Modern Track A Test set. We try to obtain a dataset out of the four datasets that help the model to do better. We chose the Faster-R-CNN resnext101_64x4d (cardinality = 64 and Bottleneck width = 4) model as the baseline model. The Mmdetection toolbox was used to implement the model with the default training configurations provided by the framework.

Evaluation metrics for ICDAR 19 dataset are based on IoU (Intersection over Union) to evaluate the performance of table region detection. Precision, Recall and, F1 scores are calculated with IoU threshold 0.6, 0.7, 0.8 and 0.9 respectively. The Weighted-Average F1 (WAvg.) is calculated by assigning a weight to each F1 value of the corresponding IoU threshold. As a result, the F1 scores with higher IoUs are given more importance than those with lower IoUs. The details of the metric are further explained by Gao *et al.* [8]. Table 1 shows the F1-scores for the IoU thresholds of baseline models on the ICDAR Test (Track A Modern). And, the model trained on the dataset having images of both augmentation techniques performs significantly better than other dataset models.

These results proved that both image transformation techniques for data augmentation help the model learn more effectively. So, we use both image transformation techniques on our General dataset for further experiments on the table detection task.

Table 1: F1-scores of the baseline models

| Dataset | IoU | | | | WAvg. |
|---|---|---|---|---|---|
| | 0.6 | 0.7 | 0.8 | 0.9 | |
| Original | 0.836 | 0.816 | 0.787 | 0.634 | 0.758 |
| Dilation | 0.869 | 0.855 | 0.835 | 0.705 | 0.807 |
| Smudge | 0.863 | 0.853 | 0.839 | 0.684 | 0.801 |
| Both | **0.888** | **0.884** | **0.863** | **0.736** | **0.835** |

To show the comparative analysis of the CascadeTabNet model with all other Cascade R-CNN and HRNet based object detection and instance segmentation models, we use the General dataset with both augmentation techniques for training. We use Mmdetection based implementation of all the models using the default configurations. All of these models have pre-trained backbones on ImageNet dataset using training schedules as of 1x (12 epochs) and 2x (24 epochs), further described in [4]. And all models utilize the Feature Pyramid Network (FPN) neck. We fine-tuned the following object detection and instance segmentation models.

Table 2: Result of models on ICDAR Test (Track A Modern)

| Model | IoU | | | | WAvg. |
|---|---|---|---|---|---|
| | 0.6 | 0.7 | 0.8 | 0.9 | |
| Retina | 0.818 | 0.785 | 0.762 | 0.664 | 0.749 |
| FRcnnHr | 0.889 | 0.877 | 0.862 | 0.781 | 0.847 |
| CRccnHr | 0.927 | 0.910 | 0.901 | 0.833 | 0.888 |
| CRcnnX | **0.929** | **0.913** | **0.903** | **0.852** | **0.895** |
| CMRcnnD | 0.912 | 0.897 | 0.880 | 0.834 | 0.877 |
| CMRcnnX | 0.931 | 0.925 | 0.909 | 0.868 | 0.905 |
| CMRcnnHr | **0.941** | **0.932** | **0.923** | **0.886** | **0.918** |

1. Retina : Resnext-101 based RetinaNet model with car-

dinality = 32 and bottleneck width = 4d.

2. FRcnnHr : Faster R-CNN with hrnetv2p_w40 backbone (40 indicates the width of the high-resolution convolution).

3. CRcnnX : Three staged Cascade R-CNN with Resnext-101 backbone having cardinality = 64 and bottleneck width = 4d.

4. CRcnnHr : Three staged Cascade R-CNN with hrnetv2p_w32 backbone.

5. CMRcnnD : Three staged Cascade R-CNN with Resnet-50 backbone with c3-c5 (adding deformable convolutions in resnet stage 3 to 5).

6. CMRcnnX : Three staged Cascade mask R-CNN with Resnext-101 backbone having cardinality = 64 and bottleneck width = 4d.

7. CMRcnnHr : Three staged Cascade mask R-CNN with hrnetv2p_w32 backbone.

Table 2 shows the evaluated F1-scores of all models on the ICDAR Test (Track A Modern) set. As seen in the table, the multi-stage cascaded network methodology along with HRNet backbone based models dominate other models. And, instance segmentation models do better than the object detection models. The Cascade mask R-CNN HRNet models achieves the highest accuracy among all models because of the fusion of two methodologies of multi-staged cascading and high-resolution convolutions used for instance segmentation.

### 5.2. Table detection evaluation

We again perform the iterative transfer learning technique to fine-tune our General model (Cascade mask R-CNN HRNet) on ICDAR 13, ICDAR 19 and TableBank datasets respectively for evaluation.

First, we fine-tune Cascade mask R-CNN HRNet on the ICDAR 19 track A train set along with dilation transform augmentation, and the following results were obtained on the modern tack A test set. We achieved 3rd rank on the post-competition leader board according to weighted-average metrics but attained the best accuracy for IoU 0.9, Table 3. The winner of the competition TableRadar performs two types of post-processing over the original output from the network. They merge the regions whose overlapped areas are larger than the defined threshold. And, detect lines in candidate table regions such that if the detected line extends over table-border, the table region is extended accordingly. The runner up NLPR PAL used Fully Convolutional Network (FCN) to classify image pixels into two categories: table and background, then table regions are extracted with Connected Component Analysis (CCA). Further details about both the datasets are described in [8]. The advantage of our approach over the approaches of the winner and runner-up is that both of these approaches involve some kind of post-processing after the original output of the network. But, in our approach, we do not perform any type of post-processing. *Our model directly outputs the accurate table region masks leveraging its architectural design and the techniques implanted during its training*.

Table 3: Comparison with participants of ICDAR 19 Track A (Modern) F1-scores [8]

| Team | IoU | | | | WAvg. |
|---|---|---|---|---|---|
| | 0.6 | 0.7 | 0.8 | 0.9 | |
| TableRadar | 0.969 | 0.957 | 0.951 | 0.897 | 0.940 |
| NLPR-PAL | 0.979 | 0.966 | 0.939 | 0.850 | 0.927 |
| Ours | **0.943** | **0.934** | **0.925** | **0.901** | **0.901** |

Evaluation metrics for TableBank dataset for table detection are based on, calculating the Precision, Recall, and F1 in the same way as in [9], where the metrics for all documents are computed by summing up the area of overlap, prediction, and ground truth. At this point, we want to emphasize that, *we only use 1500 images from word, 1500 from latex and 3000 images for word+latex datasets for training(fine-tuning) the models*. We achieved the best accuracy results for all of the three subsets, Table 4.

Table 4: TableBank results comparison with baseline[17]

| Dataset | Model | Precision | Recall | F1 |
|---|---|---|---|---|
| Both | ResNeXt-101 | 95.93 | 90.44 | 93.11 |
| | ResNeXt-152 | 96.72 | 88.95 | 92.67 |
| | Ours | **92.99** | **95.71** | **94.33** |
| Latex | ResNeXt-101 | 87.44 | 95.12 | 91.12 |
| | ResNeXt-152 | 87.20 | 96.24 | 91.49 |
| | Ours | **95.92** | **97.28** | **96.60** |
| Word | ResNeXt-101 | 95.77 | 76.10 | 84.81 |
| | ResNeXt-152 | 96.50 | 80.32 | 87.67 |
| | Ours | **94.35** | **95.49** | **94.92** |

Evaluation metrics for ICDAR 2013 is based on completeness and purity of the sub-objects of a table. We calculate precision and recall for each table and then take the average, as done by [18]. The metrics is further described by [18], [10] and [24]. *We only use 40 images from the dataset for fine-tuning the general model and 198 images for testing, while [18] and [21] used only 34 images for testing and rest of the dataset for training*. Results are shown in Table 5.

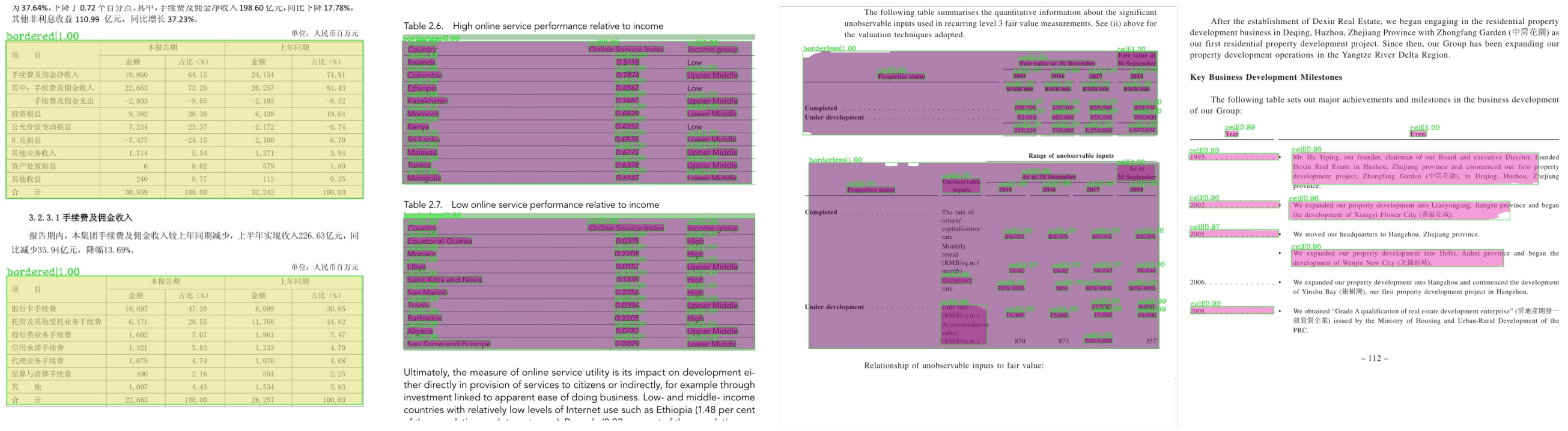

(a) Bordered table (b) Borderless table (c) Some Cells not detected (d) Table and cells not detected

Figure 5: Results of CasacadeTabNet Model

Table 5: Results of ICDAR 13 Table detection

| Model | Recall | Precision | F1-score |
|---|---|---|---|
| Ours | **1.0** | **1.0** | **1.0** |
| DeepDeSRT [21] | 0.9615 | 0.9740 | 0.9677 |
| TableNet [18] | 0.9628 | 0.9697 | 0.9662 |

## 5.3. Table structure recognition evaluation

We trained the general model on our annotated dataset, and this model is included in the final pipeline. The results are evaluated on the ICDAR 19 Track B2 dataset. The evaluation for this track is done by comparing the structure of a table that is defined as a matrix of cells. For each cell, it is required to return the coordinates of a polygon defining the convex hull of the cell's contents. Additionally, it also requires the start/end column/row information for each cell. It uses cell adjacency relation-based table structure evaluation (based on Gobel *et al*. [10]). Similar to track A, precision, recall and, F1 scores are calculated with IoU threshold of 0.6, 0.7, 0.8 and 0.9 respectively. We attain the highest accuracy on the post-competition leaderboard (Table 6), but some high-end post-processing can improve the results significantly.

Table 6: Comparison with participants of ICDAR 19 Track B2 (Modern) F1-scores [8]

| Team | IoU | | | | WAvg. |
|---|---|---|---|---|---|
| | 0.6 | 0.7 | 0.8 | 0.9 | |
| Ours | **0.438** | **0.354** | **0.190** | **0.036** | **0.232** |
| NLPR-PAL | 0.365 | 0.305 | 0.195 | 0.035 | 0.206 |

We did not use TableBank Dataset for table structure evaluation because ground truth information provided for the images only contain table structure labels in the form of HTML tags. It does not contain cell or column coordinates, and hence cannot be used to evaluate the performance of object detection or instance segmentation model. And we did not use ICDAR 13 for table structure evaluation because the evaluation metrics of ICDAR 13 uses a text content of the cell-based mapping of ground truth cells and predicted cells. For this concern, we need to extract the text content using an OCR (Optical Character Recognition) engine. And the overall accuracy would also depend on the accuracy of the OCR. We also feel that ICDAR 19 is a better metric than ICDAR 13 where the mapping of the cells is done using IoU thresholds.

Figure 5 shows the results of our model. It predicts yellow masks for bordered tables (5 a.) and purple masks for borderless tables (5 b.). It predicts accurate cell masks for most of the borderless tables. For some images where some of the predictions for cells are missed by the model (5 c.), we correct it using line estimation and contour-based text detection algorithm. The model fails badly for some images (5 d.)

## 6. Acknowledgments

We thank the following contributions because of which the paper was made possible

1. The MMdetection project team for creating the amazing framework to push the state of the art computer vision research and which enabled us to experiment and build state of the art models very easily.

2. Our college "Pune Institute of Computer Technology" for funding our research and giving us the opportunity to work and publish our research at an international conference.

3. Kai Chen [2] for endorsing our paper on the arXiv to

[2]http://chenkai.site

## 7. Conclusion

This paper presents an end-to-end system for table detection and structure recognition. It is shown that existing instance segmentation based CNN architectures which were originally trained for objects in natural scene images are also very effective for detecting tables. And, iterative transfer learning and image augmentation techniques can be used to learn efficiently from a small amount of data. The model starts learning for a general task and iteratively it learns to perform well on specific tasks. The proposed system CascadeTabNet recognizes structures within tables by predicting table cell masks while using the line information as well. Improving the post-processing modules can further enhance the accuracy of the end to end model. Our system performs better on various public datasets for both the tasks.